\documentclass[10pt,twocolumn,letterpaper]{article}

\usepackage[pagenumbers]{iccv} % To force page numbers, e.g. for an arXiv version

\definecolor{iccvblue}{rgb}{0.21,0.49,0.74}
\usepackage[pagebackref,breaklinks,colorlinks,allcolors=iccvblue]{hyperref}

\usepackage{multirow}	% for multirow
\usepackage{graphicx} 	% for rotatebox

\usepackage{amsmath}
\usepackage{amsfonts}       % blackboard math symbols
\usepackage{mathrsfs}
\usepackage{filecontents}

\def\paperID{12998} % *** Enter the Paper ID here
\def\confName{ICCV}
\def\confYear{2025}

\title{AlignDiff: Learning Physically-Grounded Camera Alignment via Diffusion}

\author{
Liuyue Xie$^1$ \quad
Jiancong Guo$^2$ \quad
Ozan Cakmakci$^2$ \quad
Andre Araujo$^3$ \\
László A. Jeni$^1$ \quad
Zhiheng Jia$^2$ \\
\\
$^1$Carnegie Mellon University \quad
$^2$Google \quad
$^3$Google DeepMind
}

\begin{document}
\maketitle
\begin{abstract}
Accurate camera calibration is a fundamental task for 3D perception, especially when dealing with real-world, in-the-wild environments where complex optical distortions are common. Existing methods often rely on pre-rectified images or calibration patterns, which limits their applicability and flexibility. In this work, we introduce a novel framework that addresses these challenges by jointly modeling camera intrinsic and extrinsic parameters using a generic ray camera model. Unlike previous approaches, AlignDiff shifts focus from semantic to geometric features, enabling more accurate modeling of local distortions. We propose AlignDiff, a diffusion model conditioned on geometric priors, enabling the simultaneous estimation of camera distortions and scene geometry. To enhance distortion prediction, we incorporate edge-aware attention, focusing the model on geometric features around image edges, rather than semantic content. Furthermore, to enhance generalizability to real-world captures, we incorporate a large database of ray-traced lenses containing over three thousand samples. This database characterizes the distortion inherent in a diverse variety of lens forms. Our experiments demonstrate that the proposed method significantly reduces the angular error of estimated ray bundles by $\sim 8.2^\circ$ and overall calibration accuracy, outperforming existing approaches on challenging, real-world datasets.

% Camera calibration is a fundamental task for 3D vision that requires estimating both intrinsic parameters (e.g. focal length, lens distortion) and extrinsic parameters (camera pose) for accurate scene understanding. However, most existing learning-based calibration methods address only part of this problem. 
% . These limitations mean they cannot fully recover a camera’s orientation and may not capture the true complexity of real-world lens distortions. In this paper, we propose a more comprehensive calibration framework that jointly estimates both intrinsic and extrinsic parameters from visual data, providing a complete solution beyond prior single-image intrinsic calibration. Importantly, our approach is trained on data grounded in real lens distortions rather than purely synthetic distortions, ensuring the learned model reflects physically meaningful optical behavior. Experiments on diverse real-world datasets demonstrate that our method achieves state-of-the-art accuracy in camera calibration, outperforming existing methods on both intrinsic and extrinsic estimation. This more holistic and physically grounded approach to calibration improves generalization across varied scenes and camera types, making it highly practical for downstream applications in 3D reconstruction, augmented reality, and robotics.
\end{abstract}    
\section{Introduction}
\label{sec:intro}

% add a teasor - intuition with three 
% example results from previous ones 
% try a gaussian splatting (maybe)

Accurate camera calibration, involving the estimation of intrinsic parameters such as focal length and lens distortions and extrinsic parameters such as camera pose, is essential for robust 3D perception in real-world environments. However, prevailing methods typically address these components independently, focusing on either intrinsic calibration or pose estimation while relying on simplified camera models that are insufficient to capture complex real-world optical aberrations. This separation is fundamentally limiting, as intrinsic and extrinsic parameters are closely interdependent; inaccuracies in modeling lens distortions propagate directly to pose estimation errors. Consequently, joint optimization of intrinsic and extrinsic parameters is critical to ensure reliable calibration under unconstrained conditions.

% Accurate camera calibration is fundamental for tasks that require precise spatial understanding, especially in real-world settings where complex optical aberrations challenge traditional models. Early advancements in camera calibration relied on predefined camera models to estimate intrinsic parameters like focal length and distortion coefficients. While effective for standard optics, this model struggles with the intricate, non-uniform aberrations found in more advanced optical systems, such as those in augmented and virtual reality. Recent learning-based methods, including Deepfocal \cite{deepfocal} and  MisCaliDet \cite{9197378}, introduced neural architectures to predict focal length and other intrinsic parameters, streamlining calibration for many practical scenarios. However, these models remain limited in their ability to capture the full spectrum of optical aberrations present in diverse real-world lenses.

\begin{figure}[t]
\centering
\includegraphics[width=0.9\linewidth, trim={0 0 0 0}]{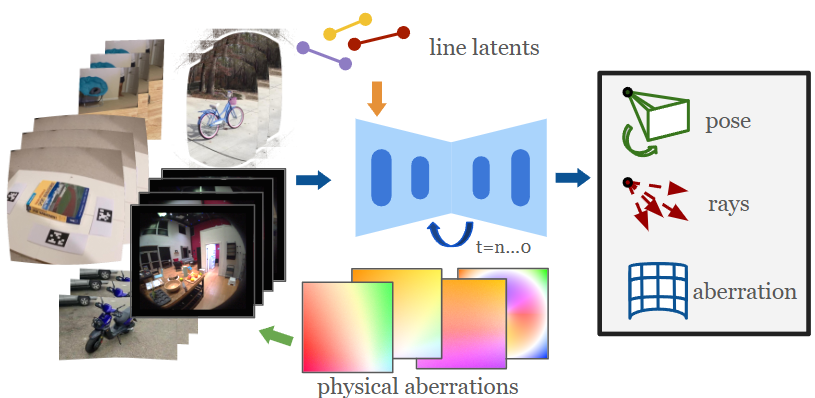}
\caption{\textbf{AlignDiff} is proposed to address common image geometric aberrations with a unified ray camera representation while jointly recovering the camera extrinsics. With groundings on physical camera lens designs, as well as the disassociation of geometric cues from semantic features, it demonstrates an ability to generalize to real video sequences.}
\vspace{-5mm}
\label{teaser}
\end{figure}

Recent diffusion-based and transformer-powered models, such as PoseDiffusion \cite{Wang2023PoseDiffusion:Adjustment}, RayDiffusion \cite{Zhang2024CamerasDiffusion}, and DiffusionSfM \cite{DiffusionSfM} have improved calibration accuracy and zero-shot generalizability by conditioning on image features extracted from vision encoders. However, these methods typically focus on high-level, semantic image features rather than on structural cues critical for modeling optical aberrations that directly impact geometry. This is particularly limiting because accurate intrinsic calibration is essential for precise extrinsic calibration: aberrations in the intrinsic parameters can significantly affect the accuracy of extrinsic parameters, making the two tightly interdependent.

\begin{figure*}[!htbp]
\centering
% \vspace{-10mm}
\includegraphics[width=\linewidth, trim={0 0 0 0}]{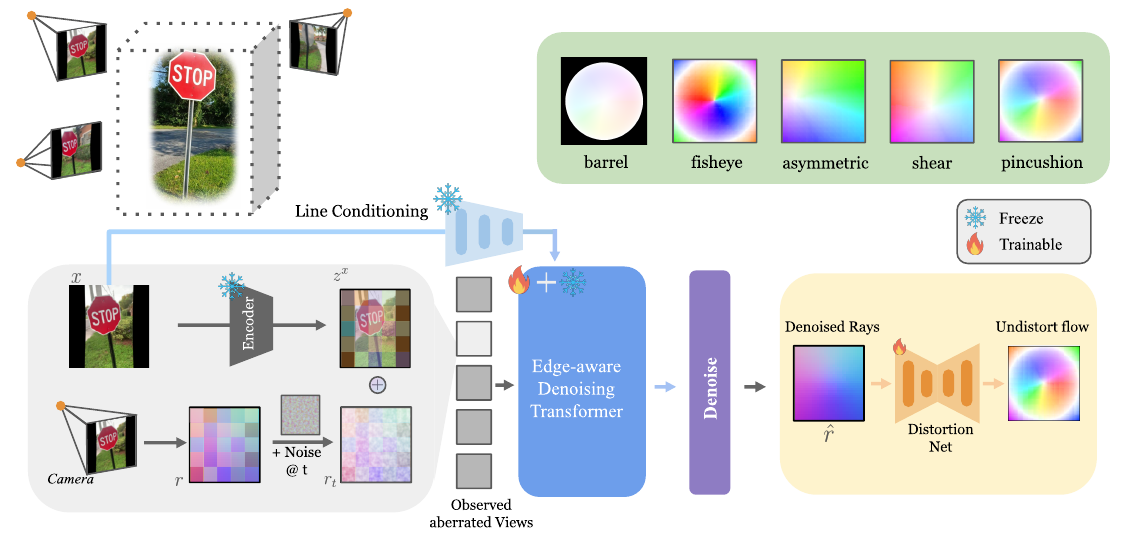}
\caption{\textbf{AlignDiff Architecture.} We promote learning camera ray profiles in three main steps: geometric cue conditioning from line features, edge-aware attention, and physical camera groundings.}
\label{network1}
\end{figure*}

We propose AlignDiff, shown Figure \ref{teaser}, a diffusion-based calibration framework, to learn the fine-grained ray profile from video sequences in world space, with a conditioning strategy that better captures fine-grained optical distortions. To shift focus from semantic to structural features, we condition the diffusion model with line embeddings from a line detection network. This approach emphasizes the geometric structures, helping the model prioritize optical aberrations over object content. In addition, we further introduce edge-focused attention to further enhance the model's sensitivity to regions around edges, where aberrations often manifest most prominently.

An additional limitation of prior models is their reliance on simulated optical aberrations derived from predefined models, which often lack the subtlety of real-world aberrations, such as localized or asymmetric profiles. To address this, our framework incorporates real optical profiles from actual lens designs, grounding the model in authentic optical characteristics and boosting its generalizability.

In summary, our framework builds upon existing diffusion-based calibration techniques by incorporating targeted conditioning strategies and real-world optical data, providing a robust solution that accurately captures complex optical aberrations. By effectively addressing intrinsic aberrations, our model lays a solid foundation for accurate extrinsic calibration, making it highly applicable for in-the-wild scenarios requiring precise camera parameters.

\begin{itemize}
    \vspace{-0.05cm}\item To our knowledge, AlignDiff is the first unified diffusion-based approach addressing complex optical aberrations in joint intrinsic and extrinsic calibration.
    \vspace{-0.05cm}\item By conditioning the diffusion model on line embeddings and incorporating edge-focused attention, our method prioritizes structural over semantic features, achieving generalization to diverse, real-world environments without requiring extensive retraining. 
    % \vspace{-0.05cm}\item We extract aberration profiles from denoised ray bundles, allowing undistorting the images with the recovered profile. 
    \vspace{-0.05cm}\item Our approach incorporates authentic optical profiles from actual lens designs, grounding the model in real-world aberrations and improving generalization to natural images.
\end{itemize}
\section{Related Works}
\label{sec:related_works}

\noindent\textbf{Aberration representation for in-the-wild images.}
Traditional camera calibration methods model intrinsic characteristics, such as focal length and camera center, alongside distortion parameters that describe image geometric aberrations. Early deep learning-based methods \cite{deepfocal, Chen2022DFNet:Matching, He2024DiffCalib:Generation, Wang2023DUSt3R:Easy, Sarlin2018FromScale, YangInnovatingArchitecture, Gentile2022LeMoN:Networks} enabled calibration from images. Recent diffusion based approaches like PoseDiffusion \cite{Wang2023PoseDiffusion:Adjustment}, extend from prior approaches with improved zero-shot performance on wild captures. 
While effective for rectified images, these methods are limited in their ability to account for local optical aberrations resulting from factors like wear, temperature fluctuations, and lens designs.

Generic ray camera models were introduced to better capture local geometric distortions \cite{Sturm2004ACalibration, Ramalingam2017ACalibration} by modeling ray bundles across image patches, offering a more detailed representation of pixel deviations compared to the pinhole model. While accurate, they require denser calibration patterns, especially when optimized using methods like PnP \cite{pnp}. Several recent work \cite{He2024DiffCalib:Generation, Zhu2023TameCalibration, Zhang2024CamerasDiffusion} adopted the ray representation in deep learning and proved its feasibility. Yet, these prior works remain largely restricted to the pinhole model and did not exploit the expressiveness of ray representation for aberration modeling. The recent blind camera undistortion line of work \cite{li2019blind, Liao2024MOWAMI} infers geometric displacements from a single image, often as a preprocessing step to provide rectified images.

Our work advances these approaches by introducing a unified ray-based framework that models both pinhole projections and local distortions. We directly estimate the ray profile and camera aberrations from a raw monocular sequence, enhancing generalizability in diverse real-world conditions.

\noindent\textbf{Joint Intrinsic and Extrinsic Calibration}
Classical camera calibration methods often assume a predefined camera model for each type of aberration, relying on calibration patterns and multi-view geometric constraints to estimate camera parameters from a sequence of images \cite{Kendall2015PoseNetAC, jin2023perspective, veicht2024geocalib, Wang2023PoseDiffusion:Adjustment, Lin2023RelPose++:Observations, kummerle2020unified, hagemann2023deep}. Early approaches solved joint intrinsic and extrinsic calibration through point matching across views \cite{guo2024vanishing, NEURIPS2019_8e6b42f1, Jiang2022Few-ViewPoses, Chen2022DFNet:Matching, Cavallari2020Real-TimeCascade, schoenberger2016sfm, pan2022camera}, with refinements summarized in \cite{liao2023deep}. Structure-from-motion (SfM) techniques later optimized reprojection loss to refine parameters further.

For in-the-wild images, Hold-Geoffroy et al. \cite{10128718} leveraged DenseNet \cite{8099726} to estimate parameters such as horizon angle and vertical field of view. Other methods \cite{jin2023perspective, 2020_jau_zhu_deepFEPE} decouple parameter estimation from direct regression, and with integrated semantic or geometric guidance \cite{JiangOmniGlue:Guidance, DeTone2017SuperPoint:Description, sarlin2020superglue, fujimoto2023deep}. Further advancements have shown that models pretrained with geometric understanding can act as better feature encoders \cite{Potje2024XFeat:Matching, 9197378, Sarlin2018FromScale,He2024DiffCalib:Generation, smith24flowmap, TrivignoTheRefinement, Yin_2018_CVPR, Jiao_2021_CVPR, Parameshwara_2022_CVPR}. Recent methods, including PoseDiffusion \cite{Wang2023PoseDiffusion:Adjustment}, DuSt3R \cite{Wang2023DUSt3R:Easy}, RayDiffusion \cite{Zhang2024CamerasDiffusion} use diffusion and transformer to recover camera parameters or ray profiles directly from images, though they still assume fixed distortion models, limiting generalization to diverse real-world aberrations \cite{liao2023deep, Liao2024MOWAMI}. The camera parameters are recovered by conditioning on local image features to capture local and cross-view geometric associations. However, these methods can yield residual errors around object contours due to reliance on semantic rather than purely geometric cues \cite{gholami2023latent, guo2023cross, 9711282}.

Our approach extends the ray representation without a fixed distortion model, disentangling semantic and geometric cues to reduce contour errors and enhance calibration accuracy. By integrating undistortion and calibration in a unified framework, we increase flexibility and applicability for real-world captures.

\section{AlignDiff}
\label{method}

\begin{figure}[!htbp]
\centering
% \vspace{-10mm}
\includegraphics[width=\linewidth, trim={0 0 0 0}]{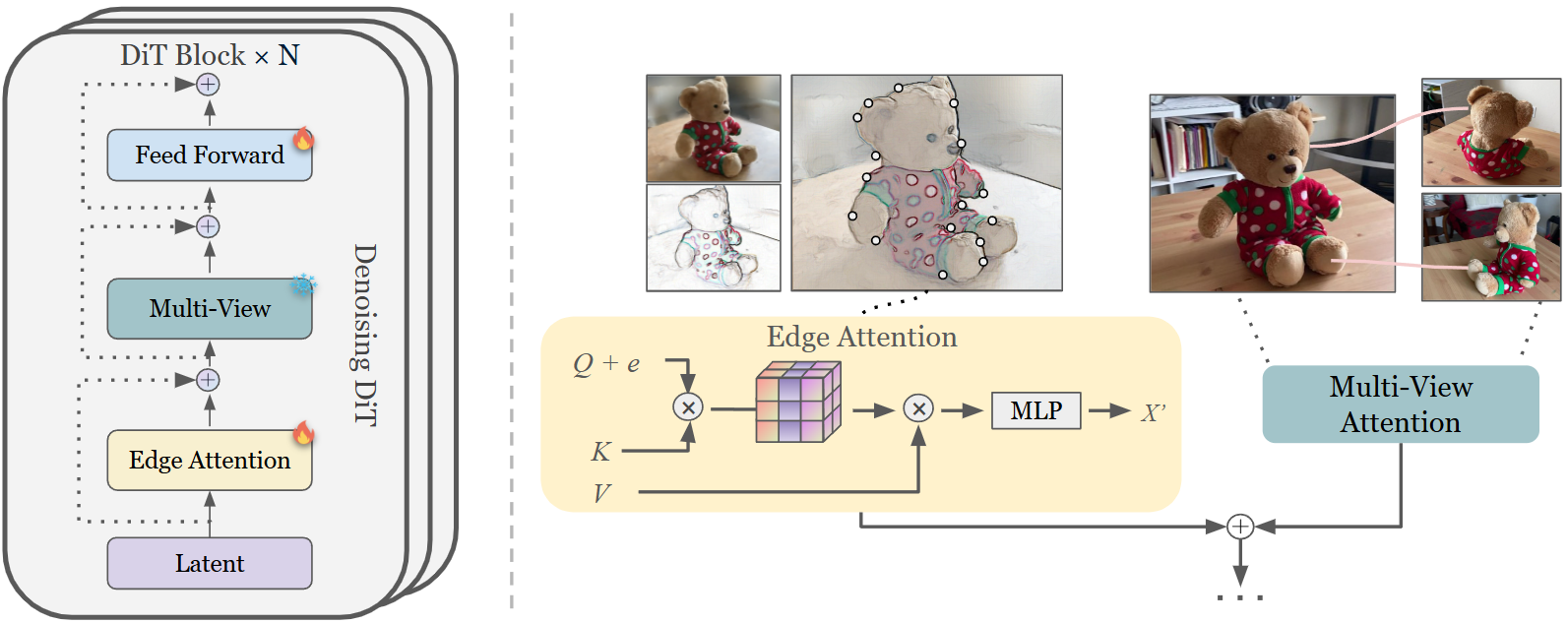}
\caption{The latents are reweighted through edge-attention that aggregates a learned mask to the Query feature to promote information along edges where the geometric cues are more prominent. Following, the multi-view attention further captures features from different views.}
\label{network2}
\end{figure}
We aim to recover generic cameras represented as ray bundles in world space from a set of $N$ input images $\{\mathcal{I}_1, \mathcal{I}_2, \mathcal{I}_3, \dots, \mathcal{I}_N\}$. Our approach, as presented in Figure \ref{network1}, captures fine-grained geometric distortions while maintaining compatibility with classical parameterized camera models. This formulation is conceptually similar to works like \cite{Zhu2023TameCalibration, Zhang2024CamerasDiffusion}, but extends the ray camera model to account for local distortions (Sec. \ref{ray_formulation}). In contrast, previous methods assume a pinhole camera model, neglecting distortions, even though they predict rays for local image patches. We adopt a Diffusion Transformer as the base architecture while enhancing this architecture by introducing geometric cues that guide ray directions based on local geometry, disentangling image perceptual features from distortions (Sec. \ref{diffuser}). From the predicted rays, we propose DistortionNet to recover the underlying lens geometric aberration that describes the ray profile's deviation from a perspective pinhole camera with the same focal length and field of view (Sec. \ref{distortion_net}). We further discuss the selected optimization objectives in Sec. \ref{loss}.

\subsection{Generic Ray Camera Representation}
\label{ray_formulation}
The objective of camera calibration is to recover optimal parameters that describe the projection of 3D world points $\mathcal{X} \in \mathbb{R}^3$ to 2D sensor locations $\boldsymbol{x} \in \mathbb{R}^2$. Our method models this with a ray-based camera model, where each unit of ray $\boldsymbol{r}_i \in \mathbb{R}^6$ is stored on a grid centered on image patches. Rays originate from the camera center $\boldsymbol{r}_o \in \mathbb{R}^3$, with directions $\boldsymbol{r}_d \in \mathbb{R}^3$ derived by lifting 2D grid points to the camera frame using: 
\begin{equation} 
    \boldsymbol{r_d} = \boldsymbol {K}^{-1}\mathcal{D}_\zeta(x), 
\end{equation} 
where $\mathcal{D}_\zeta$ is the aberration function parameterized by $\zeta$, and $\boldsymbol{K}$ is the intrinsic matrix, accounting for geometric aberrations. The ray bundle, transformed to world coordinates by rotation $\boldsymbol{R}$ and translation $\boldsymbol{t}$, \begin{equation} \boldsymbol{r}_w = \boldsymbol{R}^T \boldsymbol{r} + \boldsymbol{t}, \end{equation} represents aberrated cameras in world coordinates, allowing holistic downstream recovery of rays, camera extrinsics, and optical aberrations. 

Off-the-shelf image quality is completely characterized by the modulation transfer function (MTF) and distortion. The MTF performance includes all monochromatic and chromatic aberrations. Typically, the driving residual aberration in high-quality off-the-shelf lenses is optical distortion ($W_{311}$ in the wavefront to third order\cite{sasian2012introduction}), therefore, in this paper, our ray trace models focus on extracting the optical distortion characteristics from our dataset of lenses. 

% mention that distortions are more prominant than blurriness in the cameras
% distortion is the driving residual abberation in off-the-shelf cameras
% the blurriness measured by MTF is usually negligible?

% In contrast to WildCamera \cite{Zhu2023TameCalibration} which only infers ray directions in the local camera frame, unaware of the camera's spatial orientation, our model holistically recovers the camera's local ray profile, describing the projection mapping, as well as its extrinsics in the world frame. 
% In practice, we found that our spatial-aware framework recovers more accurate intrinsic estimations, likely because the method corrects any residual distortions from the images. 
% ray bundle/profile?
\subsection{Geometry-controlled ray diffuser}
\label{diffuser}
We train a multi-view diffusion model $\mathcal{M}_\phi$ that takes multiple images of a 3D scene as input and generates the corresponding output camera ray bundles given their geometric aberration cues. Specifically, given $N$ conditional views containing the images and their corresponding geometric aberration latents $\mathcal{Z}^g$ from a line-segment detection network, the model learns to capture the joint distribution of $N$ target ray bundles $\mathcal{R}^{tgt}=\{\mathcal{R}_1, \mathcal{R}_2, \mathcal{R}_3, \dots, \mathcal{R}_N\}^{tgt}$ from noised ray bundles $\mathcal{R}^{\epsilon}$:
\begin{equation}
    p(\mathcal{R}^{tgt} | \mathcal{I}, \mathcal{Z}^g, \mathcal{R}^{\epsilon}).
\end{equation}

\noindent\textbf{Diffusion model architecture.}
Diffusion models approximate the data manifold by learning to invert a diffusion process from data to a presumed distribution through a denoising process. Our adopted Denoising Diffusion Probabilistic Model (DDPM) specifically defines the noise distribution to be Gaussian and transitions to the next noising step:
\begin{equation}
    q(x_t | x_{t-1}) = \mathcal{N}(x_t ; \sqrt{1 - \beta_t} x_{t-1}, \beta_t \mathbb{I}), 
\end{equation}
where $\{\beta_1, \beta_2, ..., \beta_T\}$ are variances within the $\mathbf{T}$ step noising schedule, and $\mathbb{I}$ is the identity matrix. 
Similar to prior works \cite{Zhang2024CamerasDiffusion, NEURIPS2023_d6c01b02, Wang2023PoseDiffusion:Adjustment}, we adopt a DiT \cite{HoDenoisingModels} as the backbone of the diffusion model, receiving image latent embeddings conditioned on aberration cues. Given a set of sequential images and target ray bundles, the model encodes each image into a latent representation $\mathcal{Z}^c_i$  through an image feature encoder TIPS \cite{maninis2024tips}, as well as a conditioning branch comprised of multiple ResNet \cite{7780459} blocks. The image latent feature is concatenated to a noised ray bundle of the same dimension. Then, the diffusion model is trained to estimate the joint distribution of the latent-conditioned ray bundle given the conditioning geometric cues. We initialize the model from a DiT model trained for 3D shape generation, with an input resolution of $448 \times 448 \times 3$. We directly inflate the latent space of the original DiT to connect with the concatenated features of noised rays and image latents, while inheriting the rest of the trained model parameters to leverage the perceptual knowledge.

\noindent\textbf{Utilizing geometric cues.} 
The conditioning branch is designed to extract guidance information that describes geometric aberration cues and align it with the denoising features. The conditioning branch is a trained line segment detection network \cite{li2021ulsd}. We observe that the controls maintain a high level of consistency with the denoising features and eliminate the need to be inserted at multiple stages, as also indicated in \cite{Peng2024ControlNeXt:Generation}. We thus integrate the controls at a single middle block into the denoising model by adding them to the denoising features after the cross-normalization. It can serve as a plug-and-play guidance module that provides geometric aberration information about the conditioning views, such that the diffuser model is aware of the levels and types of aberration from geometric cues. 

\noindent\textbf{Edge Attention.} 
We introduce Edge Attention responsible for parsing features along the edges, as shown in Figure \ref{network2}, which highlights the edge-aware attention mechanism designed to prioritize geometric cues along image edges. In particular, edge attention takes the input embedding of an image sequence $\mathcal{Z}^c \in \mathbb{R}^{N \times H \times W \times C}$, allowing it to perform self-attention operations across the edges. We apply a patchification operator with patch size $p \times p$ to generate patch tokens $\boldsymbol{t} \in \mathbb{R}^{L\times3}$ with $L=(N\times H \times W / p^2)$ denoting the length of patchified tokens. To enable the model to disentangle the geometric information from the semantics, we integrate the edge information $e_{i...N}$ for guidance into the self-attention mechanism:
\begin{equation}
Attention(Q, K, V) = Softmax(\frac{(Q+e^T)K}{\sqrt{H}}V)
\end{equation}
where $H$ is the dimension size of each head.
The edge embedding is designed as a soft weighting to highlight the features along the geometrically prominent areas. The image edges are first extracted from the original image sequences using a Canny Edge detector. We then patchify them and apply a 3D convolution, mapping the embedding as $e_i \in \mathbb{R}^L$. 
Ultimately, we obtain the aggregated features $\hat{\mathcal{Z}_i}$, and we employ a feedforward network as in the original DiT before proceeding to the subsequent blocks. 

\subsection{Aberration profiling with DistortionNet}
\label{distortion_net}
We adopt warping flow as a unified representation to model geometric aberrations of the image compared to ideal pinhole cameras, similar to prior blind camera undistortion frameworks \cite{li2019blind, Liao2024MOWAMI}. To query the aberration profile, we connect the denoised rays to a network devised to estimate deviations of predicted ray bundles to that of a pinhole camera. The DistortNet $\mathcal{M}_\theta$, with a backbone of MAE-Tiny \cite{mae}, learns the mapping from the predicted set of ray bundles $\mathcal{R}^{tgt}$ to a set of warping flow maps $\mathcal{F}^{tgt}$. The learned flow maps can be applied to the image sequences for image undistortion. 

\begin{figure*}
\centering
% \vspace{-10mm}
\includegraphics[width=\linewidth, trim={0 0 0 0}]{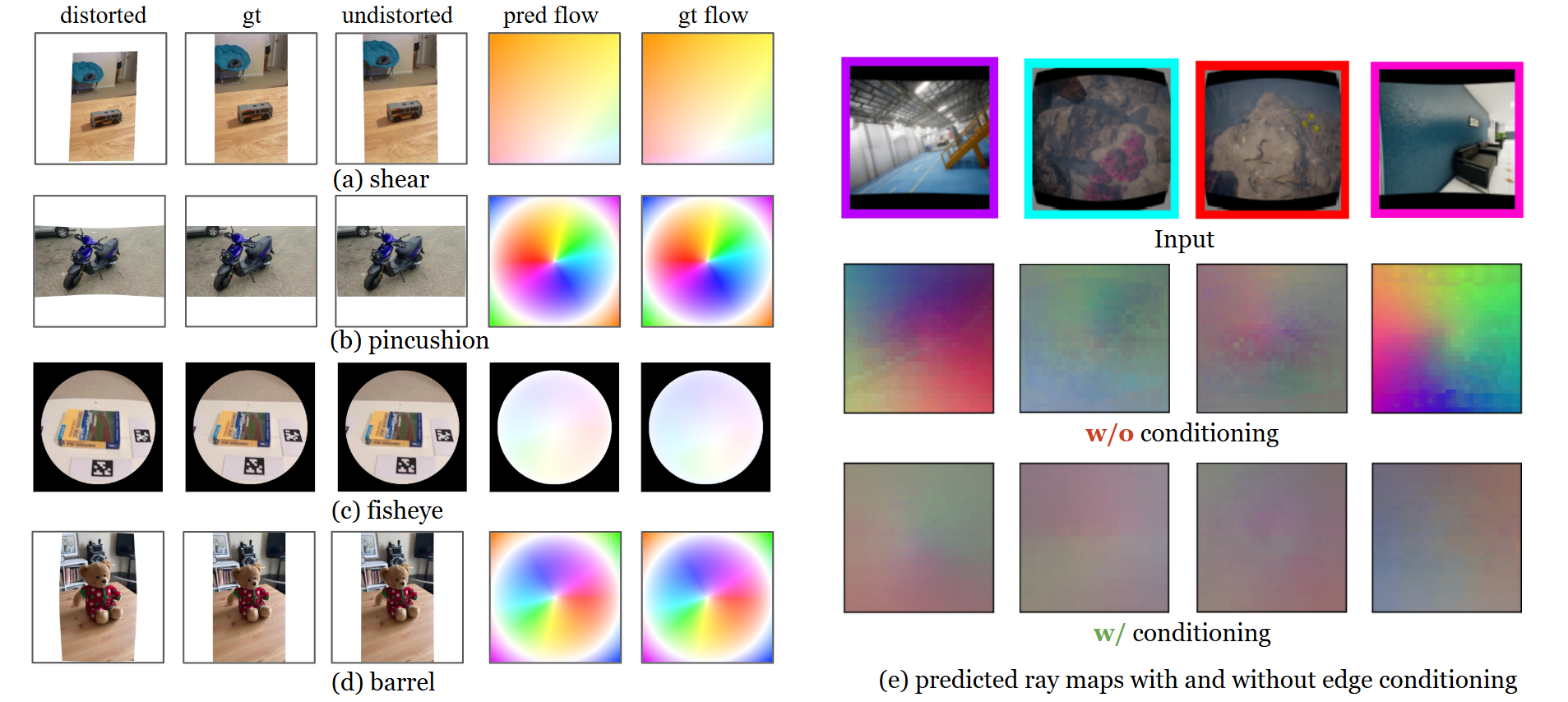}
\caption{Recovered aberration profile and undistorted images. From denoised rays in world space, the DistortNet estimates the aberration pattern represented in warping flow. The undistorted images maintain coherent structural distribution compared to aberration-free images.}
\label{undistort}
\vspace{-1em}
\end{figure*}

\subsection{Optimization objectives}
\label{loss}

% We now explain the optimization objectives for estimating the ray bundles and for extracting the underlying distortion map given $M$ images $\{I_1, I_2, I_3, ..., I_M\}$. We note that predicting distorted ray bundles from in the wild images is under-constrained and prone to uncertainties due to insufficient overlapping frustrums and possible image artifacts, eg. motion blur, lighting condition changes, etc. To handle the uncertainties arise from the capturing conditions, we train the network through a denoising process where the noised samples are reversed to recover the noise-free sample, then compared with the reference sample:
% \begin{equation}
%     \mathscr{L}_{denoise}(\phi) = \mathbb{E}_{t,  \boldsymbol{r}_o, \epsilon} ||\boldsymbol{r}_o - \mathcal{M}_\phi(\boldsymbol{r}_t, t)||^2,
% \end{equation}
% where $r_o$ is the original ray bundle, $\mathcal{M}_\phi$ is the denoising diffusion model parameterized by $\phi$, and $r_t$ is the noised sample by adding a time-dependent Gaussian noise at time $t$ to the reference sample $r_o$. 
% The noisy rays are obtained by adding Gaussian noise to the sample, and weighted by a hyperparameter $\epsilon \sim \mathcal{N}(0, \mathbb{I})$ as:
% \begin{equation}
%     \boldsymbol{r}_{t} = \sqrt{\bar{\alpha}_t} \boldsymbol{r}_o + \sqrt{1-\bar{\alpha}_t}\epsilon,
% \end{equation}
% Since with the denoising process, the training objective reverts back to a regressive loss, it enables us enforcing additional geometric constraints. 

We describe the optimization objectives for estimating ray bundles and extracting the distortion map from $N$ images $\{\mathcal{I}_1, \mathcal{I}_2, \dots, \mathcal{I}_N\}$. Estimating distorted ray bundles from in-the-wild images is under-constrained and affected by uncertainties from limited overlap and artifacts (e.g., motion blur, lighting changes). To address this, we train the network with a denoising process that reverses noised samples to match the reference: 
\begin{equation} 
    \mathscr{L}_{denoise}(\phi) = \mathbb{E}_{t,  \boldsymbol{r}_o, \epsilon} ||\boldsymbol{r}_o - \mathcal{M}_\phi(\boldsymbol{r}_t, t)||^2, 
\end{equation} 
where $\boldsymbol{r}_o$ is the original ray bundle, $\mathcal{M}_\phi$ is the denoising model, and $\boldsymbol{r}_t$ is the sample with Gaussian noise at time $t$: 
\begin{equation} 
    \boldsymbol{r}_{t} = \sqrt{\bar{\alpha}_t} \boldsymbol{r}_o + \sqrt{1-\bar{\alpha}_t}\epsilon,
\end{equation} 
where $\epsilon \sim \mathcal{N}(0, \mathbb{I})$. The denoising reverts to a regressive loss, allowing additional geometric constraints.

Recovering local camera ray profiles from an image sequence is susceptible to rotational ambiguities, leading to errors in ray bundle and pose estimates. We introduce a ray directional loss to measure angular errors in ray bundles:
% Recovering the local camera ray profiles from an image sequence is prone to rotational ambiguities due to uncertainties, which propogates towards inaccurate ray bundle estimates in the world space resulting in inaccurate pose estimates. We propose a ray directional loss measuring the angular errors in the resultant ray bundles:
\begin{equation}
    \mathscr{L}_{angular}(\phi) = cos ^{-1} ( \frac{r_o \cdot \mathcal{M}_\phi(\boldsymbol{r}_t, t)}{|r_o| |\mathcal{M}_\phi(\boldsymbol{r}_t, t)|}),
\end{equation}
DistortNet is trained to regress backward flow, representing deviations from perspective ray bundles, by comparing each predicted flow map $\boldsymbol{f}_o$ to the reference set $\mathcal{F}^{tgt}$: 
% The DistortNet is trained to regress the backward flow describing the deviation of the predicted ray bundles to their perspective counterparts. We train the network by comparing the regressed flow map to the reference flow:
\begin{equation}
    \mathscr{L}_{distort}(\theta) = \mathbb{E}_{t,  \boldsymbol{f}_o, \boldsymbol{r}_t}||\boldsymbol{f}_o - \mathcal{M}_\theta(\mathcal{M}_\phi(\boldsymbol{r}_t, t))||^2.
\end{equation}
This way, we recover the camera distortion maps at each image pixel, which can be used to directly unwarp the images and their associated rays. 

\subsection{Physical aberration groundings}
% adding the dataset descriptions from the rebuttal
% [Louise] do stats and include a short table here
% [Louise] include visulaizations of the distribitions in the appendix

\begin{figure}[!t]
\centering
% \vspace{-10mm}
\includegraphics[width=\linewidth, trim={0 0 0 0}]{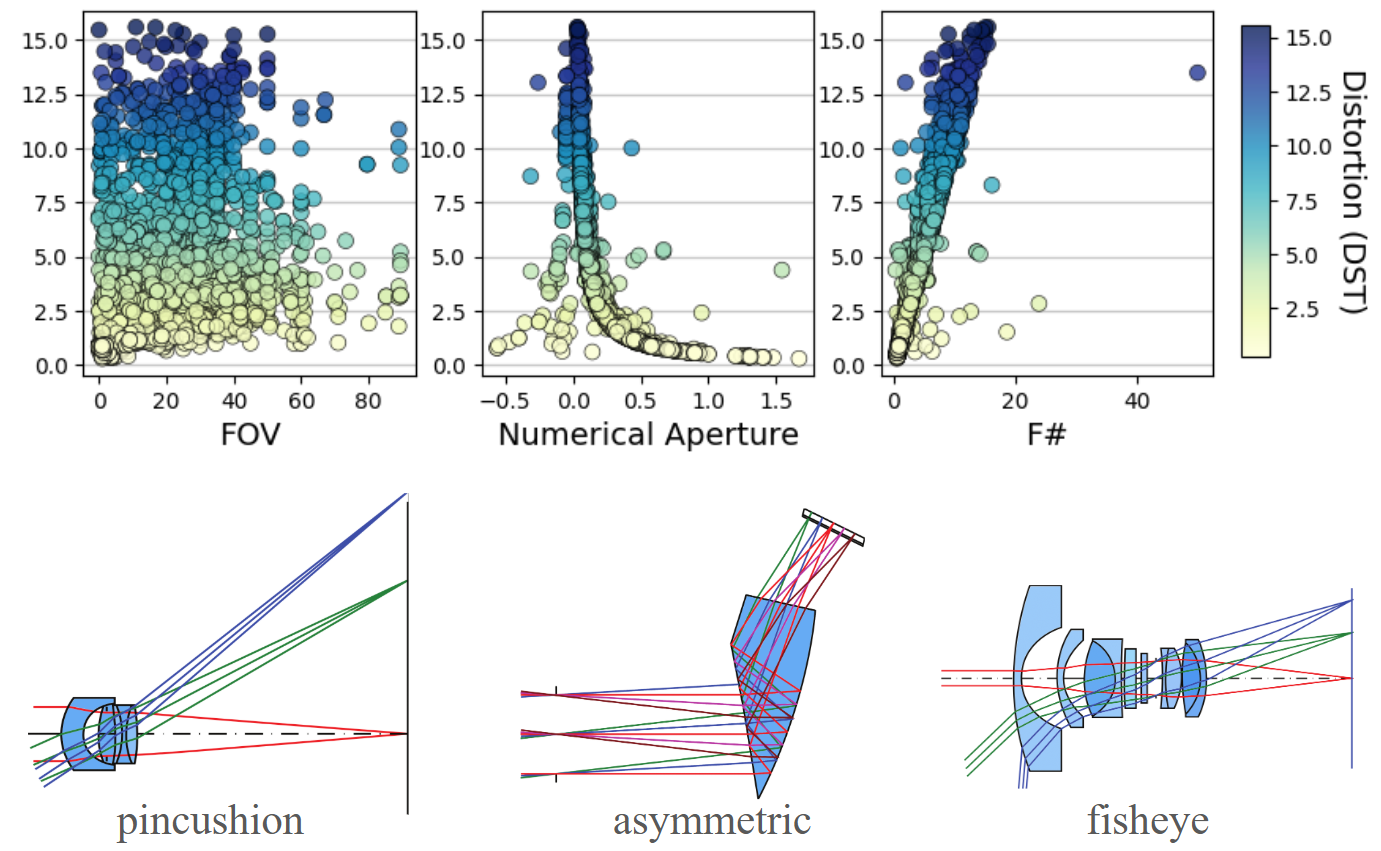}
\caption{Ray-traced lens designs are utilized, enabling the accurate simulation of geometric aberrations. The aberrations are encoded as local geometric distortion \cite{ISO17850}, with a percentage deviation of pixels from their ideal positions on a regular grid.}
\label{cam_modalities}
\end{figure}

We use the largest available lens database to ray trace and extract distortion characteristics from about 3000 optical systems. We wrote custom scripts that extracted each lens prescription from the LensView database and used commercial raytracing software to extract the optical distortion maps. These systems include camera lenses, lithography lenses, and freeform optics, as shown in Fig \ref{cam_modalities}. These systems produce a diverse set of optical distortion functions that map the object to the image. Detailed description on the lens dataset can be found in Appendix \ref{appendix_lens}. 

Since images within each dataset are usually captured with similar equipment sharing identical camera settings, to learn different camera modalities, we augment the sequences upon loading with sampled camera distortion profiles representing shear, barrel, fisheye, and pincushion. We further include a set of diverse optics dataset comprised of $3187$ patented lens profiles from LensView \cite{lensview}. These profiles span a wide range of professional and industrial lenses. Figure \ref{cam_modalities} shows examples from the grounding lenses, along with the distribution of the distortion with respect to the field of view (FOV), Numerical Aperture, and F-number. This augmentation introduces groundings in intrinsic and distortion profiles, addressing the scarcity of camera profiles within the datasets. We assess the out-of-distribution distortions not present in the lens set through experiments on the Aria Digital Twins dataset in Table \ref{eval_angle}. 

\section{Experiments}
\label{experiments}
In this section, we first provide the implementation details of the proposed framework and then validate on two real world datasets with different camera modalities. Our method outperforms baseline approaches in both aberrated and non-aberrated images, achieving state-of-the-art results in quantitative and qualitative evaluations. 

\subsection{Implementation Details}
\label{implement}
We use a pre-trained TIPS \cite{maninis2024tips} network as the image feature extractor, which gives high-quality, fine-grained image embeddings. We adopted a DiT \cite{HoDenoisingModels} as the base structure of the denoising diffuser, and augmented the attention blocks with the proposed edge conditioning attention. We train our diffusion model with $T=100$ timesteps, with the training taking roughly $3$ days on $8$ H100 GPUs.

Following prior works \cite{Wang2023PoseDiffusion:Adjustment, Lin2023RelPose++:Observations, Zhang2024CamerasDiffusion}, we use the first camera as the coordinate anchor to define the scene. The scene is rescaled such that the first camera has a unit from translation and is rotated to grant the first camera identity rotation. In this way, all the following predicted cameras would use the first camera as a reference to express their relative orientations. 

\subsection{Dataset and Evaluation Metrics}
\noindent\textbf{Datasets.} 
We chose CO3D \cite{ReizensteinCommonReconstruction} as the primary training dataset for the generation of ray maps. The dataset consists of roughly 37k  $360^\circ$ videos of common household objects from 51 MS-COCO categories. The dataset is annotated with COLMAP \cite{schoenberger2016sfm, schoenberger2016mvs} camera poses and intrinsics, with the frames undistorted with the estimated coefficients. Each video spans over on average 200 frames. The dataset provides a diverse profile of common objects sequences, making it suitable for training the network. In order to enhance the variety of training scenarios, we add MegaDepth \cite{MDLi18} and TartanAir \cite{tartanair2020iros} as our additional training data. 

Secondly, we evaluate the inference performance in Aria Digital Twins \cite{pan2023aria} that comprises 200 sequences and 400 min of videos that capture the daily activities of the common household using egocentric cameras. Its camera trajectories were provided by Optitrack \cite{optitrack} system and IMU sensor data on the Aria glass, further optimized by matching the two sources of estimated trajectories. The camera calibration parameters are provided by fitting a Kannala Brandt and Fisheye radial-tangential thin film model with the estimated trajectories. We use the same division as in \cite{pan2023aria}, with a subset of $50$ tests. With the two chosen datasets, we showcase the model's capabilities of estimating the ray maps from both common household videos and its ability to generalize to recent AR egocentric captures. 

\noindent\textbf{Evaluation protocol.}
We evaluated spare view predictions for recovered poses and the accuracy of the camera ray profile. In accordance with \cite{Zhang2024CamerasDiffusion}, we randomly sampled $N$ images for an evaluation of $N$. Each reported accuracy is averaged over $5$ runs to reduce stochasticity. 

\noindent\textit{Rotation accuracy.} We first compute the relative rotations between each pair of cameras for both predicted and ground-truth poses. The errors are computed for the pairwise relative rotations and reported for the percentage of deviations less than $15^\circ$.

\noindent\textit{Camera center accuracy.} Using the fitted ground truth and predicted poses, we align the cameras and compute the distance from the predicted camera center to the ground truth correspondence. The fraction of distances within $10\%$ of the scene scale is reported. 

\noindent\textit{Angular error.} With the recovered ray bundles, we compute the distance of the predicted rays to the ground truth correspondences. The mean angular error in degrees is reported.

\begin{figure}[!t]
\centering
\includegraphics[width=0.9\linewidth, trim={0 0 0 0}]{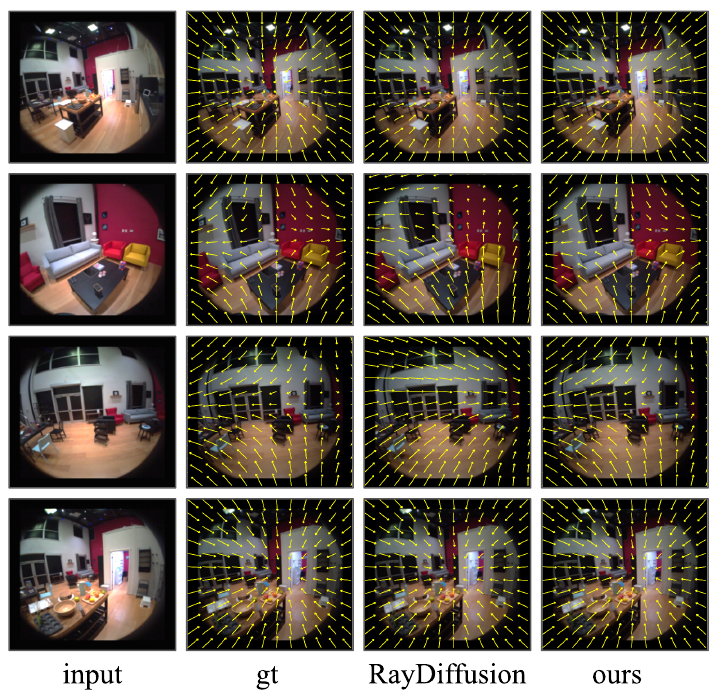}
\caption{Generalization to out-of-distribution camera from Aria Digital Twins. Our approach produces better results compared to \cite{Zhang2024CamerasDiffusion}. See supplementary for quantitative inference results.}
\label{vector}
\vspace{-1em}
\end{figure}

\subsection{Evaluation}

\begin{table}[t]
\scriptsize
\centering
\caption{\textbf{Quantitative evaluation of camera angular error} on CO3D and Aria Digital Twins datasets. Averaged angular error (degrees) across varying numbers of input images.}
\resizebox{\columnwidth}{!}{
\begin{tabular}{l c c c c c c c}
\toprule
\multicolumn{8}{c}{\textbf{CO3D – Seen Categories}} \\
\midrule
Method & 2 & 3 & 4 & 5 & 6 & 7 & 8 \\
\midrule
RayRegression \cite{Zhang2024CamerasDiffusion} & 18.4 & 20.6 & 21.7 & 21.9 & 20.3 & 23.0 & 23.7 \\
RayDiffusion \cite{Zhang2024CamerasDiffusion} & 9.6 & 9.7 & 10.2 & 8.7 & 9.4 & 11.9 & 13.2 \\
AlignDiff (Ours) & \textbf{3.4} & \textbf{3.1} & \textbf{2.8} & \textbf{3.2} & \textbf{3.0} & \textbf{3.6} & \textbf{4.1} \\
\midrule
\multicolumn{8}{c}{\textbf{CO3D – Unseen Categories}} \\
\midrule
RayRegression \cite{Zhang2024CamerasDiffusion} & 24.1 & 22.5 & 21.6 & 24.8 & 23.9 & 25.1 & 26.2 \\
RayDiffusion \cite{Zhang2024CamerasDiffusion} & 11.4 & 10.8 & 14.1 & 12.3 & 15.8 & 16.6 & 19.4 \\
AlignDiff (Ours) & \textbf{4.6} & \textbf{5.2} & \textbf{5.8} & \textbf{5.7} & \textbf{6.4} & \textbf{6.8} & \textbf{8.2} \\
\midrule
\multicolumn{8}{c}{\textbf{Aria Digital Twins}} \\
\midrule
RayRegression \cite{Zhang2024CamerasDiffusion} & 34.6 & 38.2 & 36.5 & 40.8 & 41.2 & 40.9 & 43.7 \\
RayDiffusion \cite{Zhang2024CamerasDiffusion} & 28.4 & 29.1 & 34.7 & 37.2 & 40.4 & 42.3 & 45.6 \\
AlignDiff (Ours) & \textbf{15.4} & \textbf{13.8} & \textbf{14.2} & \textbf{20.6} & \textbf{21.3} & \textbf{24.6} & \textbf{24.9} \\
\bottomrule
\end{tabular}
}
\label{eval_angle}
\end{table}

\begin{table}[t]
% \begin{tabular*}{\linewidth}{@{\extracolsep{\fill}} cccccc }
\small
\centering
\caption{\textbf{Camera Rotation and Camera Center Accuracy} on Distorted CO3D dataset. Experiments using geometrically aberrated videos, comparing recent methods against AlignDiff.}
\resizebox{0.5\textwidth}{!}{\begin{tabular}{ cc ccccccc} 
\toprule
	 \multicolumn{9}{c}{Rotation @ $15^\circ$} \\
	\hline \hline
	& number of Images  & 2 & 3 & 4 & 5 & 6 & 7 & 8 \\
	\hline
	& RelPose \cite{relpose}& 62.0 & 24.0 & 34.0 & 24.0 & 25.0 & 33.0 & 32.0\\
	& RelPose++ \cite{Lin2023RelPose++:Observations}& 72.5 & 73.4 & 73.6 & 74.7 & 75.0 & 76.0 & 75.7\\
	& PoseDiffusion \cite{Wang2023PoseDiffusion:Adjustment}& 74.5 & 74.9 & 74.4 & 74.7 & 75.1 & 75.4 & 76.0\\
	& RayDiffusion \cite{Zhang2024CamerasDiffusion}& 74.0 & 80.0 & 85.3 & 82.4 & 84.8 & 82.5&86.6\\
	\multirow{-5}{*}{\rotatebox[origin=c]{90}{Seen Cls}} & AlignDiff (Ours) & $\boldsymbol{90.6}$ &$\boldsymbol{91.1}$&$\boldsymbol{91.3}$&$\boldsymbol{91.8}$&$\boldsymbol{92.4}$&$\boldsymbol{92.7}$&$\boldsymbol{93.1}$\\
	\hline
	& RelPose \cite{relpose}&  61.0 & 27.0 & 37.0 & 26.0 & 30.0 & 26.0 & 25.0\\
	& RelPose++ \cite{Lin2023RelPose++:Observations}& 61.4 & 60.8 & 63.0 & 64.1 & 65.7 & 65.7 & 65.4\\
	& PoseDiffsuion \cite{Wang2023PoseDiffusion:Adjustment}& 74.5 & 74.9 & 74.4 & 74.7 & 75.1 & 75.4 & 76.0\\
	& RayDiffusion \cite{Zhang2024CamerasDiffusion}&  69.9 &72.4 & 75.3&76.2&76.4&77.1&78.5\\
	\multirow{-5}{*}{\rotatebox[origin=c]{90}{Unseen Cls}} & AlignDiff (Ours) & $\boldsymbol{83.6}$&$\boldsymbol{84.8}$&$\boldsymbol{84.9}$&$\boldsymbol{85.2}$&$\boldsymbol{85.7}$&$\boldsymbol{86.1}$&$\boldsymbol{86.1}$\\

	\midrule
	\multicolumn{9}{c}{Camera Center @ $0.1$} \\
    \hline \hline
	& RelPose++ \cite{Lin2023RelPose++:Observations}& 100 & 85.2 & 79.0 & 74.3 & 70.5 & 68.6 & 66.0\\
	& PoseDiffusion \cite{Wang2023PoseDiffusion:Adjustment}&  100 & 72.3 & 56.7 & 53.6 & 52.4 & 57.1 & 52.1\\
	& RayDiffusion \cite{Zhang2024CamerasDiffusion}& 100 & 73.4 & 72.5 & 71.1 & 71.0 & 70.6 & 70.2\\
	\multirow{-4}{*}{\rotatebox[origin=c]{90}{Seen Cls}} & AlignDiff (Ours) & 100 & $\boldsymbol{92.5}$& $\boldsymbol{90.7}$ & $\boldsymbol{88.6}$ & $\boldsymbol{87.2}$ & $\boldsymbol{85.8}$ & $\boldsymbol{84.6}$\\
	\hline
	& RelPose++ \cite{Lin2023RelPose++:Observations}& 100 & 66.4 & 57.0 & 51.3 & 48.9 & 44.5 & 44.1\\
	& PoseDiffusion \cite{Wang2023PoseDiffusion:Adjustment}& 100 & 72.3 & 56.7 & 53.6 & 52.4 & 57.1 & 52.1\\
	& RayDiffusion \cite{Zhang2024CamerasDiffusion}& 100 & 74.8 & 71.6 & 70.2 & 67.3 & 66.7 & 62.5\\
	\multirow{-4}{*}{\rotatebox[origin=c]{90}{Unseen Cls}} & AlignDiff (Ours) & 100& $\boldsymbol{86.3}$ & $\boldsymbol{84.2}$ & $\boldsymbol{80.5}$ &$\boldsymbol{75.2}$ & $\boldsymbol{73.0}$ & $\boldsymbol{70.1}$\\
\bottomrule
\end{tabular}}
\label{eval_distort}
\vspace{-1em}
\end{table}
\begin{table}[t]
% \begin{tabular*}{\linewidth}{@{\extracolsep{\fill}} cccccc }
\small
\centering
\caption{\textbf{Camera Rotation and Camera Center Accuracy} on Rectified CO3D dataset. AlignDiff identifies subtle local distortions, achieving improvements over previous methods.}
\resizebox{\columnwidth}{!}{\begin{tabular}{cc ccccccc} 
\toprule
	 \multicolumn{9}{c}{Rotation @ $15^\circ$} \\
	\hline \hline
	& number of Images  & 2 & 3 & 4 & 5 & 6 & 7 & 8 \\
	\hline
	& COLMAP \cite{SchonbergerStructure-from-MotionRevisited}& 30.7 & 28.4 & 26.5 & 26.8 & 27.0 & 28.1 & 30.6\\
	& RelPose \cite{relpose}& 56.0 & 56.5 & 57.0 & 57.2 & 57.2 & 57.3 & 57.2\\
	& PoseDiffusion \cite{Wang2023PoseDiffusion:Adjustment}& 75.7 & 76.4 & 76.8 & 77.4 & 78.0 & 78.7 & 78.8\\
	& RelPose++ \cite{Lin2023RelPose++:Observations}& 81.8 & 82.8 & 84.1 & 84.7 & 84.9 & 85.3 & 85.5\\
	& RayRegression \cite{Zhang2024CamerasDiffusion}& 88.8 & 88.7 & 88.7 & 89.0 & 89.4 & 89.3 & 89.2\\
	& RayDiffusion \cite{Zhang2024CamerasDiffusion}& 91.8 & $\boldsymbol{92.4}$ & 92.6 & 92.9 & 93.1 & 93.3 & 93.3\\
	\multirow{-7}{*}{\rotatebox[origin=c]{90}{Seen Cls}} & AlignDiff (Ours) & $\boldsymbol{92.6}$ & 92.3 & $\boldsymbol{92.7}$ & $\boldsymbol{93.0}$ & $\boldsymbol{93.6}$ & $\boldsymbol{93.4}$ & $\boldsymbol{93.5}$ \\
	\hline 
	& COLMAP \cite{SchonbergerStructure-from-MotionRevisited}&  34.5 &31.8 &31.0 &31.7 &32.7 &35.0 &38.5\\
	& RelPose \cite{relpose}&  48.6& 47.5& 48.1& 48.3& 48.4& 48.4& 48.3\\
	& PoseDiffusion \cite{Wang2023PoseDiffusion:Adjustment}& 63.2& 64.2 &64.2& 65.7& 66.2 &67.0 &67.7\\
	& RelPose++ \cite{Lin2023RelPose++:Observations}& 69.8 &71.1& 71.9& 72.8 &73.8 &74.4& 74.9\\
	& RayRegression \cite{Zhang2024CamerasDiffusion}&79.0& 79.6& 80.6& 81.4& 81.3& 81.9 &81.9\\
	& RayDiffusion \cite{Zhang2024CamerasDiffusion}&  83.5 & 85.6 & 86.3 & 86.9 & 87.2 & 87.5 & 88.1\\
	\multirow{-7}{*}{\rotatebox[origin=c]{90}{Unseen Cls}} & AlignDiff (Ours) & $\boldsymbol{86.3}$ & $\boldsymbol{86.6}$ & $\boldsymbol{87.4}$ & $\boldsymbol{87.7}$ & $\boldsymbol{88.5}$ & $\boldsymbol{88.7}$ & $\boldsymbol{89.2}$\\

	\midrule
	\multicolumn{9}{c}{Camera Center @ $0.1$} \\
    \hline \hline
	& COLMAP \cite{SchonbergerStructure-from-MotionRevisited}& 100 &34.5 &23.8 &18.9& 15.6& 14.5& 15.0\\
	& RelPose \cite{relpose}& 100 &76.5& 66.9 &62.4& 59.4 &58.0 &56.5\\
	& PoseDiffusion \cite{Wang2023PoseDiffusion:Adjustment}& 100& 77.5 &69.7& 65.9& 63.7& 62.8 &61.9\\
	& RelPose++ \cite{Lin2023RelPose++:Observations}& 100& 85.0 &78.0& 74.2 &71.9& 70.3& 68.8\\
	& RayRegression \cite{Zhang2024CamerasDiffusion}& 100& 91.7 &85.7 &82.1 &79.8 &77.9 &76.2\\
	& RayDiffusion \cite{Zhang2024CamerasDiffusion}&  100 &$\boldsymbol{94.2}$ &90.5& $\boldsymbol{87.8}$ &86.2 &85.0& $\boldsymbol{84.1}$\\
	\multirow{-7}{*}{\rotatebox[origin=c]{90}{Seen Cls}} & AlignDiff (Ours) &  100 & 93.7 & $\boldsymbol{91.8}$ & 87.5 & $\boldsymbol{87.1}$ & $\boldsymbol{85.2}$ & 83.9  \\
	\hline
	& COLMAP \cite{SchonbergerStructure-from-MotionRevisited}&  100 & 36.0 & 25.5 & 20.0 & 17.9 & 17.6 & 19.1\\
	& PoseDiffusion \cite{Wang2023PoseDiffusion:Adjustment}& 100& 63.6 &50.5 &45.7& 43.0 & 41.2 & 39.9\\
	& RelPose++ \cite{Lin2023RelPose++:Observations}& 100& 70.6& 58.8& 53.4 &50.4 & 47.8 & 46.6\\
	& RayRegression \cite{Zhang2024CamerasDiffusion}&  100 &83.7& 75.6 &70.8 &67.4 &65.3 &63.9\\
	& RayDiffusion \cite{Zhang2024CamerasDiffusion}&  100 &87.7 &$\boldsymbol{81.1}$& 77.0& 74.1& 72.4 &$\boldsymbol{71.4}$\\
	\multirow{-7}{*}{\rotatebox[origin=c]{90}{Unseen Cls}} & AlignDiff (Ours) & 100 & $\boldsymbol{88.2}$ & 81.0 & $\boldsymbol{79.3}$ & $\boldsymbol{77.5}$ & $\boldsymbol{72.6}$ & 71.0\\
\bottomrule
\end{tabular}}
\label{eval_rect}
\end{table}

\begin{table}[t]
\small
\centering
\caption{\textbf{Camera Rotation and Camera Center Accuracy} on Aria Digital Twins dataset.}
\resizebox{\columnwidth}{!}{
\begin{tabular}{cc ccccccc} 
\toprule
\multicolumn{9}{c}{Rotation @ $15^\circ$} \\
\hline \hline
& Number of Images & 2 & 3 & 4 & 5 & 6 & 7 & 8 \\
\hline
& RelPose \cite{relpose} & 59.6 & 32.7 & 38.0 & 31.2 & 28.6 & 27.4 & 26.0 \\
& RelPose++ \cite{Lin2023RelPose++:Observations} & 62.8 & 61.3 & 62.6 & 62.8 & 63.2 & 64.9 & 64.8 \\
& PoseDiffusion \cite{Wang2023PoseDiffusion:Adjustment} & 75.4 & 74.8 & 74.1 & 75.8 & 75.2 & 75.9 & 76.2 \\
& RayDiffusion \cite{Zhang2024CamerasDiffusion} & 75.2 & 76.3 & 78.9 & 79.2 & 79.6 & 80.0 & 80.2 \\
\multirow{-5}{*}{\rotatebox[origin=c]{90}{Unseen Cls}} & AlignDiff (Ours) & \textbf{81.4} & \textbf{82.7} & \textbf{82.8} & \textbf{84.3} & \textbf{84.1} & \textbf{84.6} & \textbf{85.5} \\
\midrule
\multicolumn{9}{c}{Camera Center @ $0.1$} \\
\hline \hline
& RelPose++ \cite{Lin2023RelPose++:Observations} & 100 & 70.3 & 56.1 & 52.8 & 48.6 & 46.4 & 44.7 \\
& PoseDiffusion \cite{Wang2023PoseDiffusion:Adjustment} & 100 & 71.8 & 57.4 & 54.6 & 51.8 & 52.2 & 48.5 \\
& RayDiffusion \cite{Zhang2024CamerasDiffusion} & 100 & 75.9 & 69.2 & 61.8 & 60.4 & 55.6 & 51.1 \\
\multirow{-4}{*}{\rotatebox[origin=c]{90}{Unseen Cls}} & AlignDiff (Ours) & \textbf{100} & \textbf{78.3} & \textbf{70.0} & \textbf{64.4} & \textbf{62.7} & \textbf{58.1} & \textbf{52.9} \\
\bottomrule
\end{tabular}
}
\label{aria_eval_rect}
\vspace{-1em}
\end{table}

% % TODO: blow the image up
% \begin{figure*}[!htbp]
% \centering
% \begin{minipage}[b]{.5\textwidth}
% \centering
% % \vspace{-10mm}
% \includegraphics[width=\linewidth, trim={0 0 0 0}]{sec/images/pose.png}
% \caption{Recovered cameras from aberrated image sequences. Predicted and expected cameras are visualized in solid and dashed lines, respectively. The cameras align with the rotational orientation of the expected counterparts. The ray moment and residual errors show low level of deviation from reference rays.}
% \label{pose}
% \end{minipage}\qquad
% \begin{minipage}[b]{.45\textwidth}
% \centering
% % \vspace{-10mm}
% \includegraphics[width=\linewidth, trim={0 0 0 0}]{sec/images/ablation.png}
% \caption{Evaluation of the proposed conditioning method in residual errors. Without conditioning, the image patterns are visible in the error plot. While the errors are more uniformly distributed and lower in magnitude with our method.}
% \label{ablate_vis}
% \end{minipage}
% \end{figure*}

\noindent\textbf{Baseline comparisons.}
In Table \ref{eval_angle}, we present the mean ray angular error measurements for the distorted CO3D dataset, demonstrating the effectiveness of our approach compared to existing methods. Our results indicate that our method excels at extracting precise, arbitrary ray profiles even in the absence of a prior camera model. This capability is particularly significant given the challenges of distorted sequences. Notably, using the same ray representation as RayDiffusion, our approach consistently surpasses its performance, where the diffusion model depends solely on image-based features. The experiments reveal that our method achieves a more robust disentanglement from local image content, effectively isolating structural cues from object-specific features. This disentanglement minimizes residual errors tied to object appearance, allowing structural details to be more accurately emphasized.

Additionally, we report metrics for camera rotation accuracy and camera center accuracy, with aberrated images in Table \ref{eval_distort} and rectified images in Table \ref{eval_rect}. Our approach consistently outperforms existing methods in estimating both camera position and orientation under aberrated conditions, confirming its resilience to challenging visual distortions. For rectified images, our conditioning method does not simply maintain performance but rather enhances it by recognizing and compensating for residual aberrations. This capability allows our approach to improve predictions even on images with minimal distortion, achieving a high level of precision and adaptability in handling diverse image qualities.
% \noindent\textbf{Recovering pose and aberration profile.}

\begin{figure}[!t]
\centering
% \vspace{-10mm}
\includegraphics[width=\columnwidth, trim={0 0 0 0}]{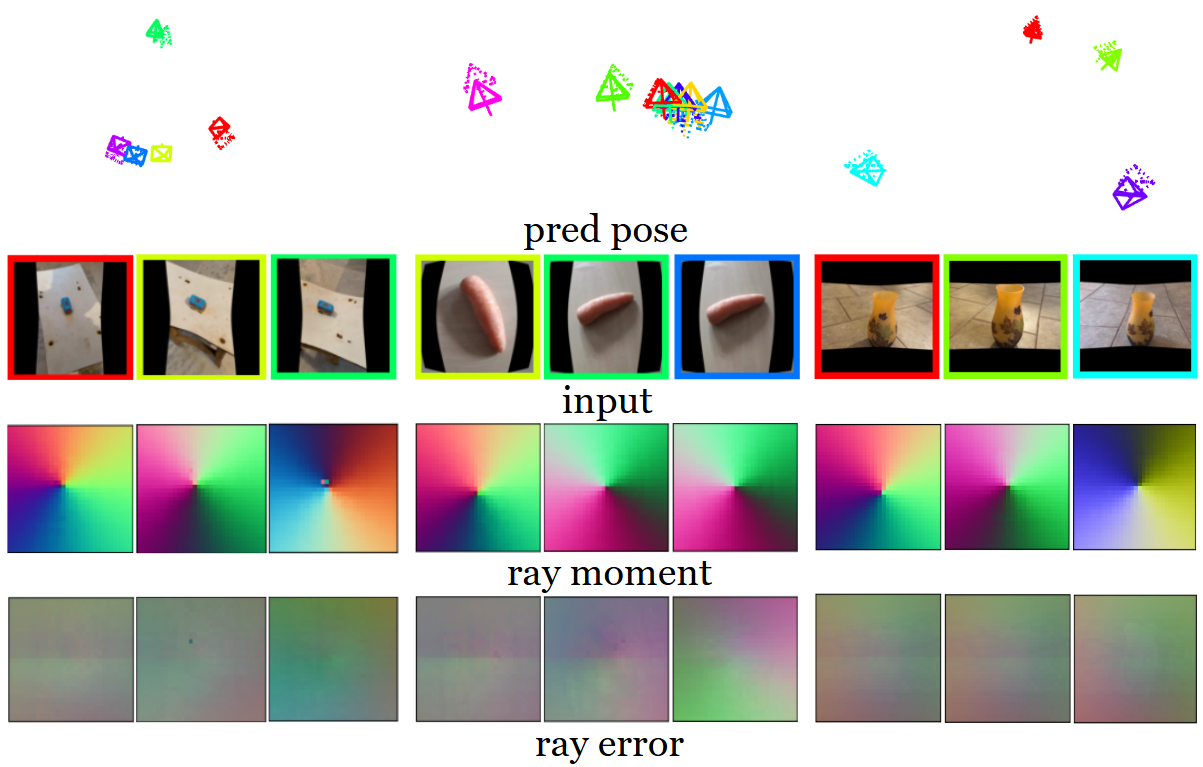}
\caption{Recovered cameras from aberrated image sequences. Predicted and expected cameras are visualized in solid and dashed lines, respectively. The cameras align with the rotational orientation of the expected counterparts. The ray moment and residual errors show low levels of deviation from reference rays.}
\label{pose}
\vspace{-1em}
\end{figure}

\noindent\textbf{Qualitative analysis of recovered cameras.} We compare the generated cameras with the ground truth cameras in world space, as shown in Figure \ref{pose}. The predicted ray moments and angular error from ground truth ray directions are also shown. Remarkably, the generated cameras exhibit little rotational difference from the reference cameras. While there are observed to have noticeable translation errors, we argue that this is due to scale ambiguities of the captures. The visualized ray moments present a smooth profile with very low angular errors. We further compare the Aria Digital Twins dataset for the inferred rays in vector form, shown in Figure \ref{vector}. These further demonstrate that our approach can achieve robust and accurate ray profiles through camera conditioning. Detailed quantitative evaluation on the Aria Digital Twins dataset are included in Appendix \ref{appendix_aria}.

\noindent\textbf{Undistortion using predicted flow.} We conduct image undistortion with the predicted aberration map, as shown in Figure \ref{undistort}. Since we adopt a unified representation for aberrations, the radial fisheye camera model's vignetting effect leads to mapping artifacts. The predicted aberration map overall correctly captures the image distortions and reliably remaps the images regardless of the aberration types.

\begin{table}[t]
\large 
\centering
\caption{Ablated framework design evaluated with averaged ray angular error.}
\vspace{-2mm}
\resizebox{\columnwidth}{!}{  % Adjust the table to fit the column width
\begin{tabular}{l | ccccccc} 
\toprule
    Methods  & 2 & 3 & 4 & 5 & 6 & 7 & 8 \\
    \hline
    Baseline & 8.8 & 9.5 & 10.4 & 9.2 & 12.4 & 12.7 & 13.9 \\
    AlignDiff (w/o angular loss) & 5.8 & 6.1 & 8.8 & 7.3 & 9.6 & 11.3 & 14.7 \\
    AlignDiff (w/o line conditioning) & 6.7 & 8.4 & 9.8 & 9.9 & 10.3 & 10.1 & 12.8 \\
    AlignDiff (w/o edge-attentions) & 5.3 & 7.1 & 7.4 & 9.2 & 10.8 & 11.3 & 10.5\\
    AlignDiff (w/o optics grounding) & 5.2 & 6.9 & 6.8 & 7.7 & 8.4 & 9.1 & 11.3\\
    AlignDiff & $\boldsymbol{4.6}$ & $\boldsymbol{5.2}$ & $\boldsymbol{5.8}$ & $\boldsymbol{5.7}$ & $\boldsymbol{6.4}$ & $\boldsymbol{6.8}$ & $\boldsymbol{8.2}$\\
\bottomrule
\end{tabular}}
\label{eval_ablate}
\end{table}

\begin{table}[t]
% \begin{tabular*}{\linewidth}{@{\extracolsep{\fill}} cccccc }
\tiny
\centering
\caption{Evaluations on different image latent encoder choices.}
\vspace{-2mm}
\resizebox{0.5\textwidth}{!}{\begin{tabular}{l | ccccccc} 
\toprule
	Methods  & 2 & 3 & 4 & 5 & 6 & 7 & 8 \\
	\hline
	DINOv2-s/14 \cite{oquab2023dinov2}& 8.2 & 8.1 & 8.6 & 10.4 & 9.8 & 10.1 & 12.5 \\
	DINOv2-g/14 \cite{oquab2023dinov2}& $\boldsymbol{4.5}$ & 6.4 & 6.3 & 6.6 & $\boldsymbol{6.1}$ & 7.8 & 9.6\\
	TIPS-g/14 \cite{maninis2024tips}& 4.6 & $\boldsymbol{5.2}$ & $\boldsymbol{5.8}$ & $\boldsymbol{5.7}$ & 6.4 & $\boldsymbol{6.8}$ & $\boldsymbol{8.2}$\\
\bottomrule
\end{tabular}}
\label{eval_vit}
% \vspace{-5mm}
\end{table}

\noindent\textbf{Ablation study.}
We perform an ablation study on CO3D data to evaluate the impact of each conditioning module, enhanced geometric guidance, and additional losses, as shown in Table \ref{eval_ablate}. For each comparison, we remove one conditioning module from the framework. The results show that all modules significantly improve performance, and latent conditioning of the line segment has the greatest effect. Figure \ref{undistort} visualizes the ray residuals with and without aberration conditioning, further validating its effectiveness. Additionally, we test the angular loss, finding that our localized ray angular loss yields notable improvements.

While our multiview diffusion approach implicitly incorporates temporal consistency by explicit temporal losses, we chose to handle temporal consistency as a post-processing step to keep the model lightweight and directly applicable in real-world deployments. This approach reduces computational overhead while still providing meaningful consistency improvements, leaving more complex temporal modeling for future work. This is based on the premise that camera profiles remain generally consistent over short periods, barring changes due to factors like camera sensor temperature fluctuations. To test this, we implemented a post-processing step that rejects low-confidence pose estimations, which reduced the average ray angular error from 8.20 ° to 8.06 ° over 8 frames. Furthermore, enforcing temporal consistency on local camera ray profiles using reprojection mean squared error loss yielded an average reduction of 0.06° in angular error across the frames. We plan to integrate these enhancements, along with outlier rejection and a frame selection mechanism, into our future work.

\noindent\textbf{Evaluations on image representation models.}
The image sequence encoder is designed to generate dense global embeddings that capture perceptual, geometric, and semantic representations. Among the available models, DINOv2 and TIPS are well suited for extracting high-quality dense features. We conduct an ablation study to evaluate these models' effectiveness in querying camera ray profiles. As shown in Table \ref{eval_vit}, the TIPS-g/14 model achieves the lowest angular errors on CO3D, with DINOv2-g/14 as a close competitor.
\section{Conclusion}
We introduce AlignDiff, a unified calibration framework for real-world optical aberrations, enabling the recovery of camera ray bundles, extrinsics, and aberration profiles from multi-view image sequences using simple conditioning techniques. First, we propose structural conditioning to separate image geometry from semantic features, promoting emphasis on structural cues for accurate ray profile estimation. Second, we aggregate features along edges, ensuring that high-quality image embeddings are paired with effective guidance to avoid semantic interference. Finally, we incorporate real-world lens profiles into training, grounding AlignDiff in actual camera designs. These contributions collectively enhance generalization and accuracy across diverse distortion profiles, as demonstrated by consistent improvements over strong baselines.

While AlignDiff performs well in controlled indoor settings, ray estimation may become unstable in arbitrary outdoor scenes due to scale ambiguity. To enhance robustness in such cases, integrating geometric priors like optical flow or combining traditional camera parameter estimation with network inference may be promising for future directions.

\clearpage
{
    \small
    \bibliographystyle{ieeenat_fullname}
    \bibliography{main}
}

% WARNING: do not forget to delete the supplementary pages from your submission 
\clearpage
\setcounter{page}{1}
\appendix

\setcounter{section}{0}

\renewcommand\thesection{\Alph{section}}
\renewcommand\thesubsection{\thesection.\roman{subsection}}
\maketitlesupplementary

\renewcommand{\thefigure}{A\arabic{figure}}
\setcounter{figure}{0}
\setcounter{table}{0}
\renewcommand{\thetable}{A\arabic{table}}

\section{Grounding lenses}
\label{appendix_lens}
\begin{figure*}[!htbp]
\centering
% \vspace{-10mm}
\includegraphics[width=\linewidth, trim={0 0 0 0}]{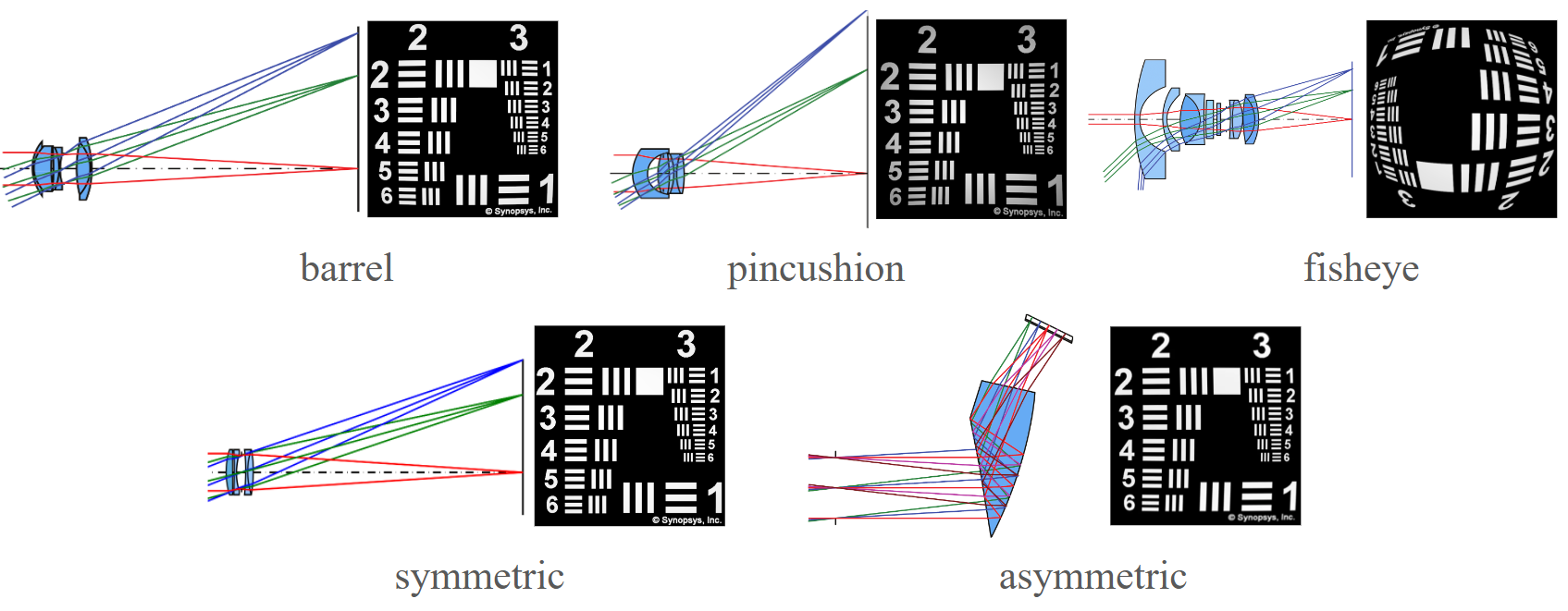}
\caption{\textbf{Selected lens designs.} The lens dataset contains more than $3000$ patented lens designs. The lens designs can be roughly categorized to resemble five aberration profiles. The geometric aberrations are converted into grid displacements, then applied to training sequences for simulation. }
\label{lens_ray}
\end{figure*}

Lenses are designed with diverse characteristics, such as curvature, surface shapes, and defining parameters, to meet specific use cases. Compound lenses, widely adopted for objectives such as achieving predefined fields of view, enhanced resolution, or improved color intensity, consist of multiple surfaces with varying materials, coatings, and geometries. Commercially available lenses exhibit significant variation in design and aberration profiles, each tailored to distinct objectives. Common geometric profiles include barrel, pincushion, fisheye, symmetric, and asymmetric designs, as illustrated in Figure \ref{lens_ray}. Barrel and fisheye aberrations cause image points to appear closer to the center compared to a uniform reference grid, while pincushion aberrations push points outward relative to the grid.

\begin{figure*}[!htbp]
\centering
% \vspace{-10mm}
\includegraphics[width=\linewidth, trim={0 0 0 0}]{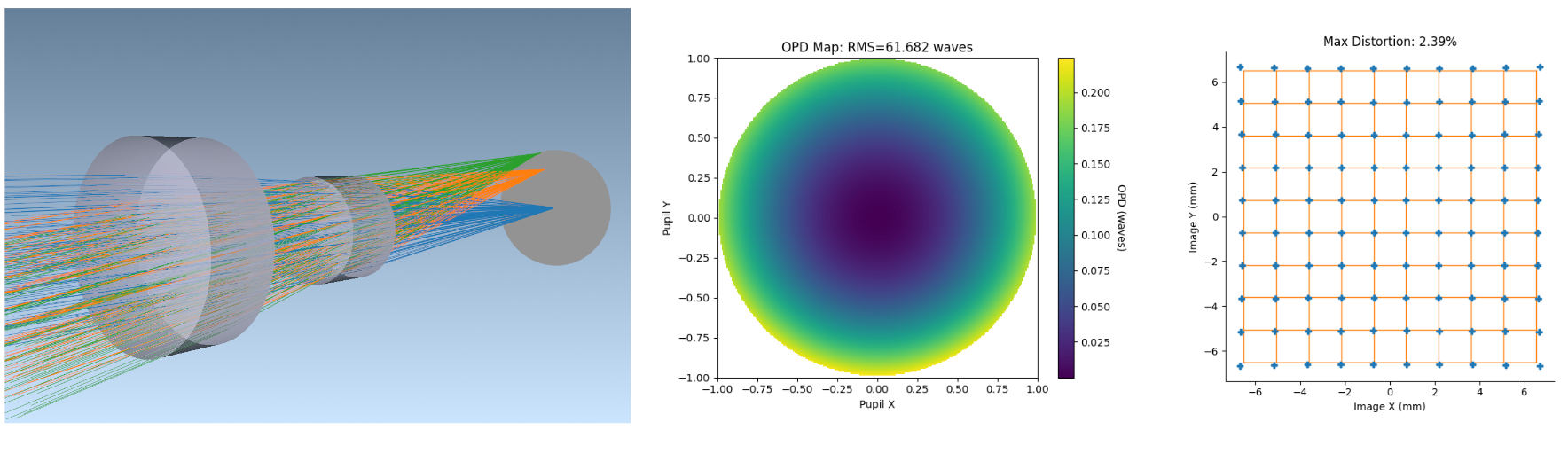}
\caption{\textbf{Ray-traced lens.} For each lens design in the simulation data, we trace rays towards the image plane to derive the aberration representations. The reversed ray profiles originating from the image plane is to be estimated through our model. The pupil diagram and displacement grid are visualized, describing the ray-traced geometric aberrations. }
\label{ray_trace}
\end{figure*}

To model lens profiles accurately, optical systems rely on ray tracing based on Snell's Law and paraxial optics. We show an example of a ray-traced optical system with geometric aberration in Figure \ref{ray_trace}. This approach derives spatially varying point spread functions (PSFs), relative illumination maps, distortion fields, and incident ray profiles. While spatially varying PSFs and illumination maps simulate perceptual aberrations on a scene image $\mathcal{I}s$, applying distortion fields alone offers a straightforward approach to simulate geometric aberrations. These aberrations are typically quantified as:

\begin{equation}
    \mathcal{D} (\%) = \frac{d_{ad} - d_{ref}}{d_{ref}},
\end{equation}
where $d_{ad}$ and $d_{ref}$ denote the actual and reference distances from the image center, respectively, with $d_{ref}$ derived via monochromatic paraxial ray tracing.

To compute distorted coordinates from an aberration profile, we perform bilinear interpolation on the distortion field at each pixel’s location on the image plane.

\begin{figure}[!htbp]
\centering
% \vspace{-10mm}
\includegraphics[width=\linewidth, trim={0 0 0 0}]{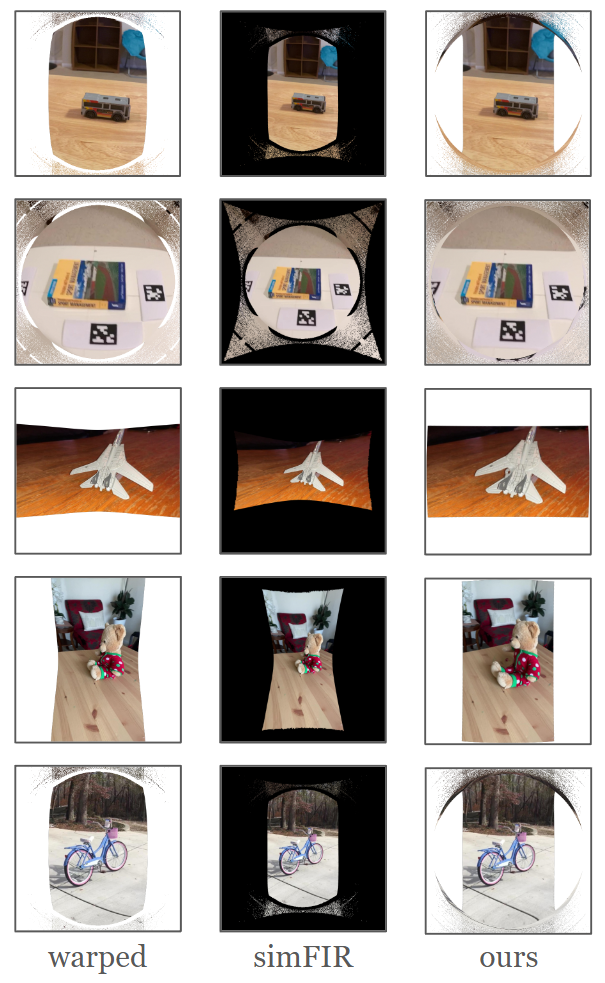}
\caption{\textbf{Aberration correction compared to existing method.} SimFIR \cite{feng2023simfir} is a recent framework for blind image undistortion. Here we showcase the correction result from our approach and SimFIR. }
\label{undistort_simfir}
\end{figure}

\section{Camera Uncertainty.}
\label{uncertainty}
\begin{figure*}[!htbp]
\centering
% \vspace{-10mm}
\includegraphics[width=\linewidth, trim={0 0 0 0}]{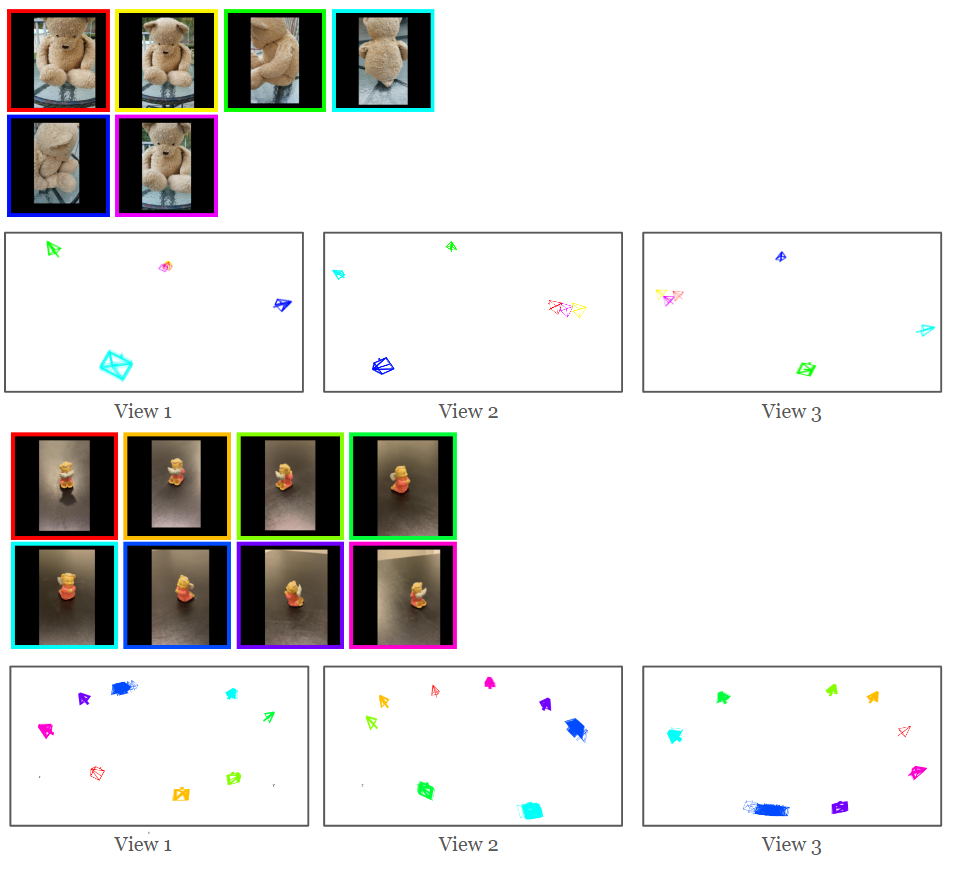}
\caption{\textbf{Uncertainty Plot.} As a diffusion model, our framework makes prediction on the cameras in a probabilistic manner. The predictions are stochastic, with each set of predicted camera sequence resembling a valid ray bundle set. }
\label{uncertainty_plot}
\end{figure*}

We present a visualization of aggregated results from our model across $20$ runs for the same input sequence in Figure \ref{uncertainty_plot}. Predicted poses are color-coded according to their respective input images. Sparse-view camera estimation inherently involves non-determinism due to scale ambiguity and symmetry. Despite these challenges, our method generates reasonable camera sets for each sequence. The predicted camera sets exhibit probabilistic variation, with larger variances observed at frames that have reflection-symmetric counterparts within the sequence.

\section{Further details on baseline finetuning}
\label{appendix_finetune}
In our comparisons, we chose PoseDiffusion, RelPose, RelPose++, RayDiffusion, and RayRegression as the data-driven baselines. The released checkpoints from these methods are not initially trained with geometrically aberrated images, leading to limited generalization for our proposed unified camera calibration task. We therefore finetune each model on the CO3D dataset using the same geometric aberration simulation as our framework for fair evaluations. For each of the model, we initialize the fine-tuning with their respective publically released checkpoint. The dataloaders are modified to include the aberration simulations, where a random aberration is selected from the lens database, and then each image in the sequence goes through the same aberration as if each video is captured using the same camera model. We finetune for $10,000$ steps, and with a learning rate of $0.00005$ for all models such that the losses are converged with the introduced camera modalities, without drifting far from the initial convergence state. 

\section{Aria Experiments}
\label{appendix_aria}
The Aria datasets are captured using geometrically aberrated fisheye cameras. Here, we test our framework for its reconstruction quality using the estimated camera pose and aberration profile. 

% \noindent\textbf{Zero-shot results.} We evaluate our model on the Aria Digital Twin dataset, assessing ray angular error and camera pose errors. As shown in Tables \ref{aria_eval_angle} and \ref{aria_eval_rect}, our predicted camera poses outperform those from previous approaches. However, accuracy is reduced compared to the CO3D dataset. This drop in performance is likely due to the unique capturing setup of the Digital Twin dataset, which captures daily egocentric activities, including dynamic hand movements not accounted for in our model's training pipeline. Incorporating proper masking for these dynamic regions could potentially improve prediction accuracy.\\
Gaussian Splatting has emerged as a common 3D reconstruction framework, conditioned on properly undistorted sequential images and camera poses. It works as a solid testing ground to check the validity of our estimated cameras both for their spatial orientations and the aberration profiles. We attempted to reconstruct with $4$ object-centric videos from the Aria Digital Twin Catalog dataset. The reconstruction results are shown in Figure \ref{render}, with corresponding videos showcased in the supplementary website. 

\begin{figure*}[!htbp]
\centering
% \vspace{-10mm}
\includegraphics[width=\linewidth, trim={0 0 0 0}]{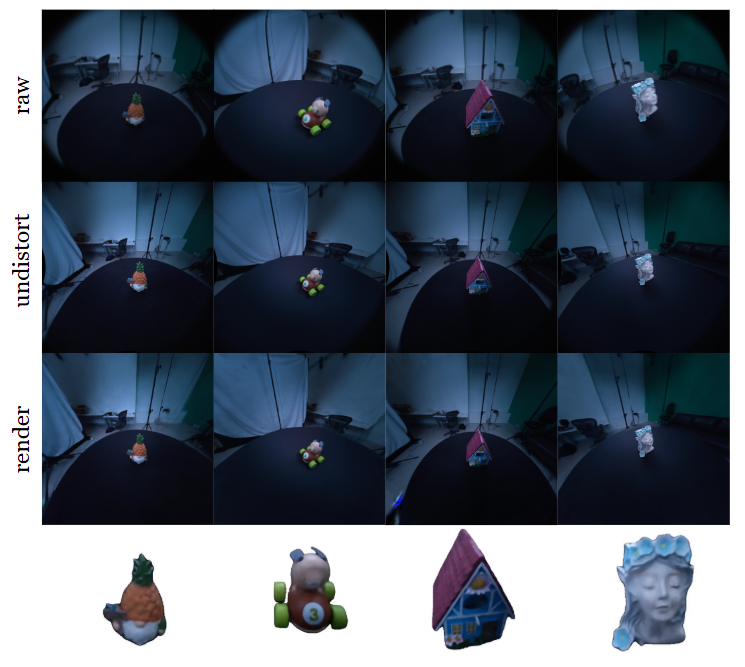}
\caption{\textbf{Gaussian Splatting Reconstruction.} Using the estimated ray bundles and distortion profiles, we process raw fisheye captures from the Aria Digital Twin Catelog dataset and train a Gaussian Splatting model for reconstruction.}
\label{render}
\end{figure*}

\section{Undistortion comparisons}
\label{appendix_undistort}
Previous approaches have explored blind image undistortion from a single monocular image. While our method performs better when applied to a sequence of images, it can be adapted for single-image inference. We compare our undistortion results against the recent SimFIR method on the barrel, fisheye, and pincushion distortions, as illustrated in Figure \ref{undistort_simfir}. SimFIR employs a tailored representation for these distortions and predicts a vignetting mask for each image. Our method generally achieves improved undistortion quality, while the vignetting masks from SimFIR help mitigate visual artifacts near image boundaries.

\clearpage

\end{document}